%% file: main.tex
\pdfoutput=1

\documentclass[11pt]{article}

\usepackage[]{acl}

\usepackage{tikz}
\usepackage{times}
\usepackage{latexsym}
\usepackage{bm}
\usepackage{amsmath}
\usepackage{amssymb}
\usepackage{amsfonts}
\usepackage{booktabs}
\usepackage{graphicx}
\usepackage{colortbl}
\usepackage{subfigure}
\usepackage{xcolor}
\usepackage{arydshln}
\usepackage{soul}

\newcommand{\hlc}[2][yellow]{{%
    \colorlet{foo}{#1}%
    \sethlcolor{foo}\hl{#2}}%
}
\DeclareMathOperator*{\argmax}{arg\,max}

\usepackage[T1]{fontenc}

\usepackage[utf8]{inputenc}

\usepackage{microtype}

\title{Efficient Argument Structure Extraction \\with Transfer Learning and Active Learning}

\author{Xinyu Hua \\
  Bloomberg \\
  New York, NY \\
  \texttt{xhua22@bloomberg.net} \\\And
  Lu Wang \\
  Computer Science and Engineering \\
  University of Michigan \\
  Ann Arbor, MI \\
  \texttt{wangluxy@umich.edu} \\
}

\begin{document}
\maketitle
\input{00abstract}

\section{Introduction}
\label{sec:intro}
\input{01introduction}

\section{Related Work}
\label{sec:related}
\input{02related.tex}

\section{Argument Relation Prediction Model}
\label{sec:model}
\input{03model}

\section{Active Learning Strategies}
\label{sec:al}
\input{04active}

\section{Datasets and Domains}
\label{sec:data}
\input{05data}

\section{Experiments and Results}
\label{sec:exp}
\input{06experiment}

\section{Conclusion}
\label{sec:conclusion}
\input{07conclusion}

\section*{Acknowledgements}
\input{08acknowledge}

\bibliography{reference}
\bibliographystyle{acl_natbib}

\newpage
\appendix
\input{09appendix}

\end{document}

%% file: 00abstract.tex
\begin{abstract}
The automation of extracting argument structures faces a pair of challenges on (1) encoding long-term contexts to facilitate comprehensive understanding, and (2) improving data efficiency since constructing high-quality argument structures is time-consuming. 
In this work, we propose a novel context-aware Transformer-based argument structure prediction model which, on five different domains, significantly outperforms models that rely on features or only encode limited contexts. 
To tackle the difficulty of data annotation, we examine two complementary methods: (i) \textit{transfer learning} to leverage existing annotated data to boost model performance in a new target domain, and (ii) \textit{active learning} to strategically identify a small amount of samples for annotation. We further propose model-independent sample acquisition strategies, which can be generalized to diverse domains. 
With extensive experiments, we show that our simple-yet-effective acquisition strategies yield competitive results against three strong comparisons. Combined with transfer learning, substantial F1 score boost (5-25) can be further achieved during the early iterations of active learning across domains.

\end{abstract}

%% file: 01introduction.tex
\begin{figure}[t]
\centering
    \includegraphics[width=80mm]{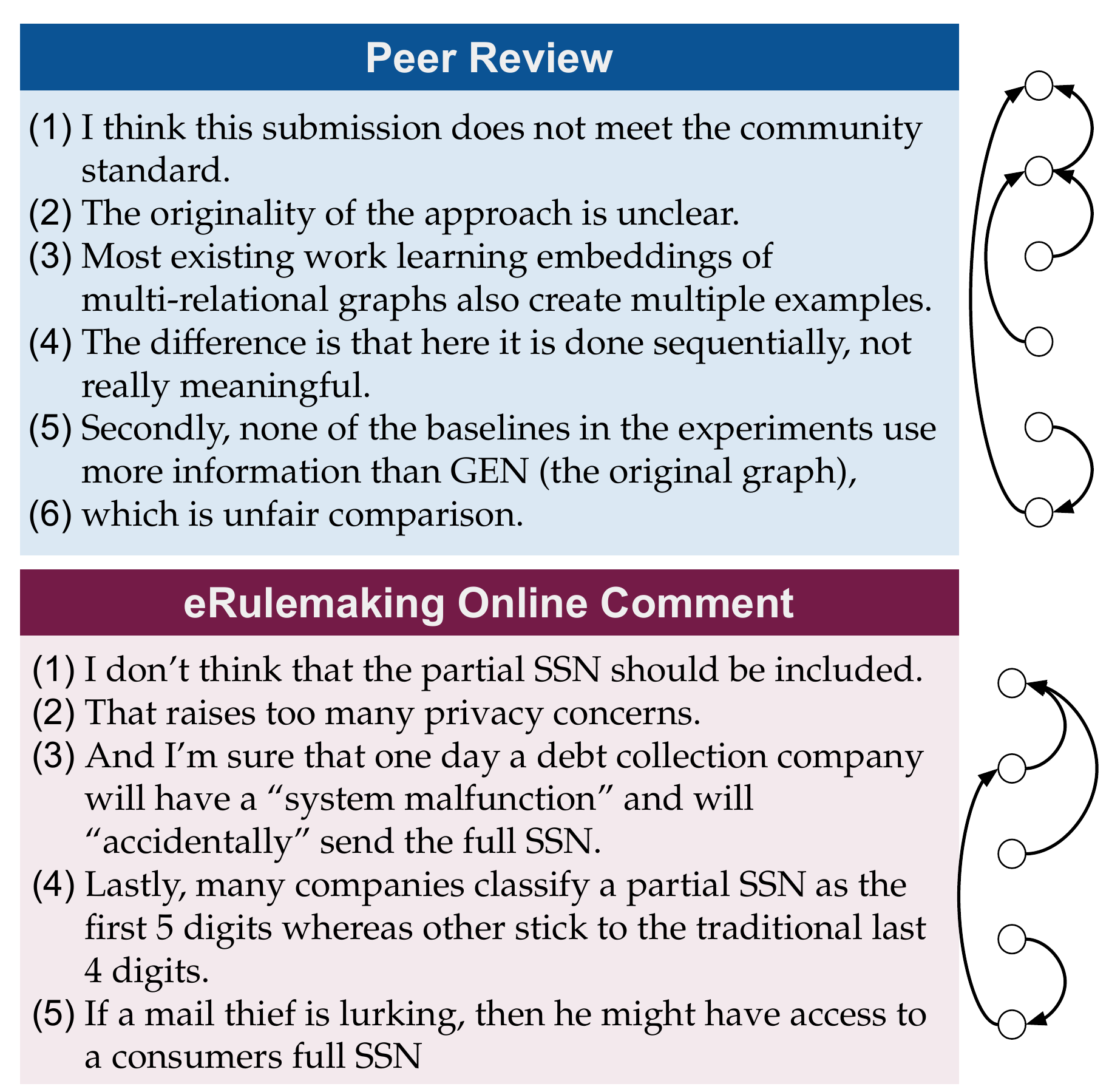}
    \caption{
    Excerpts of arguments in peer reviews and online comments. On the right, argumentative structure is labeled as support relations among propositions. 
    Despite differences in topics and vocabularies, we see similar structural patterns with long-term dependencies, motivating learning transferable representations across domains.
    }
    \label{fig:intro-example}
\end{figure}

Identifying and understanding the argumentative discourse structure in text has been a critical task in argument mining~\cite{peldszus2013argument,cabrio2018five,lawrence-reed-2019-argument,li-etal-2020-exploring-role}. It plays an important role of discovering the central theses and reasoning process across a wide spectrum of domains, from formal text such as legal documents~\cite{palau2009argumentation,lippi2016argumentation,poudyal-etal-2020-echr} and scientific literature~\cite{mayer2020ecai,fergadis-etal-2021-argumentation,al-khatib-etal-2021-argument}, to online posts and discussions~\cite{cardie-etal-2008-erulemaking,boltuzic-snajder-2014-back,park-cardie-2014-identifying,habernal-gurevych-2017-argumentation,hua2017understanding}. 
Here we focus on automatic \textbf{argumentative relation prediction}---given any proposition in a document, predict the existence and polarity (support or attack) of relation from any other proposition within a specified context window. One major challenge resides in capturing \textit{long-term dependencies}. As illustrated in Fig.~\ref{fig:intro-example}, propositions with an argumentative relation are often separated by a large text span, requiring the understanding of a longer context~\cite{nguyen-litman-2016-context,opitz-frank-2019-dissecting}.

Existing methods for this important task are often time-consuming, as they require at least three steps
~\cite{nguyen-litman-2016-context,stab-gurevych-2017-parsing,niculae-etal-2017-argument,mayer2020ecai}: (1) acquiring high-quality labels from domain experts, (2) manually designing customized features to address long dependencies and encode task-specific language, and (3) model training.
To exacerbate the challenge, the resulting models are hardly generalizable to new domains.

Consequently, our main goal is to design an \textit{easy-to-use} framework that can facilitate researchers and practitioners to build argument structure extraction models for \textit{new domains} \textit{rapidly} and \textit{accurately}. 
To this end, we first propose a novel \textit{context-aware} argument relation prediction model, which can be directly fine-tuned from pre-trained Transformers~\cite{NIPS2017_3f5ee243,liu2019roberta}. 
For a given proposition, the model encodes a broad context of neighboring propositions in the same document, and predicts whether each of them supports, attacks, or has no relation to the original one. By contrast, prior work only encodes pairwise propositions while ignoring contexts~\cite{mayer2020ecai}.

Moreover, while training on a large labeled corpus has become the \textit{de facto} method for neural models, labeling argument structures is a laborious process even for experienced annotators with domain knowledge~\cite{green-2014-towards,saint2018two,lippi2016argumentation}. 
Our second goal is to investigate \textit{efficient model training}, by using fewer samples for a new domain. We study the following two complementary techniques: 
(i) \textbf{Transfer learning} (TL) adapts models trained on existing annotated data in a different domain, or leverages unlabeled in-domain data for better representation learning.
(ii) \textbf{Active learning} (AL) strategically selects \textit{samples in the new domain} based on a sample acquisition strategy with the goal of optimizing training performance. This process is often done in multiple rounds within a given budget~\cite{settles2009active}. 
As pointed out by~\newcite{lowell-etal-2019-practical}, model-specific selection methods may not generalize across successor models and domains. 
We thus design model-independent strategies to encourage the inclusion of \textit{unseen words}, and sentences with \textit{discourse markers}. Both are easy to implement and incur little computation cost.  
We compare them with popular methods based on uncertainty~\cite{lewis1994sequential,houlsby2011bayesian} and sample diversity~\cite{sener2018active}.

For experiments, we release {\bf AMPERE++}\footnote{Data and code are available at \url{https://xinyuhua.github.io/Resources/acl22/}.}, the first dataset in the peer review domain labeled with argument relations.
Our annotation process involves over 10 months of training and multi-round sessions with experienced annotators, finally yielding $3,636$ relations over $400$ reviews originally collected in our prior work~\cite{hua-etal-2019-argument}. 
It has the highest overall relation density and the most attack relations, compared to prior datasets (Table~\ref{tab:stats}). 
We also evaluate on four other datasets covering diverse topics, including Essays~\cite{stab-gurevych-2017-parsing}, AbstRCT~\cite{mayer2020ecai} for biomedical paper abstracts, ECHR~\cite{poudyal-etal-2020-echr} for case-law documents, and the Cornell eRulemaking Corpus (CDCP)~\cite{PARK18.679} for online comments on public policies.
Our second data contribution comprises three large collections of unlabeled samples tailored for self-supervised pretraining for the first three domains.

Drawing from extensive experiment results, we make the following observations:
(1) Our proposed model, which can encode longer contexts, yields better argument relation prediction results than comparisons or variants that operate over limited contexts (\S\ref{sec:results-supervised}). 
(2) TL substantially improves performance for target domains when less labeled data is available. For example, for ECHR and CDCP, using AMPERE++ as the source domain, with only half of the target domain training data, the model achieves better F1 scores than non-transferred model trained over the entire training set (\S\ref{sec:results-tl}). This also highlights the value of our AMPERE++ data.  
(3) Among AL methods, our newly proposed model-independent acquisition strategies yield competitive results against comparisons that require significantly more computations (\S\ref{sec:results-al}). 
(4) TL further improves all AL setups {and narrows the gaps among strategies} (\S\ref{sec:results-al}).

%% file: 02related.tex
\paragraph{Argument Structure Extraction.}
Analyzing argumentation in natural language text has seen rapid growth
~\cite{lippi2016argumentation,cabrio2018five,lawrence-reed-2019-argument}, yet the most challenging aspect of it is to extract the structures among diverse argument components. 
Conceptually, the structure extraction model needs to address two subtasks: (1) determining which propositions are targeted (head detection), and (2) identifying the argumentative relations towards the head propositions. 
Early work~\cite{peldszus2013argument,peldszus-stede-2015-joint} takes inspiration from discourse parsing. While practically argument relations can be dispersed across the text, contrary to assumptions in common discourse theory~\cite{mann1988rhetorical,webber2019penn}.
More recent work considers all pairwise combinations of propositions~\cite{stab-gurevych-2014-identifying,niculae-etal-2017-argument,mayer2020ecai}, which incurs expensive computations for long documents. Our model encodes a sequence of propositions and extract their labels in one forward pass, leading to much reduced training and inference complexity while allowing access to more contexts.

\paragraph{Transfer Learning for Structured Prediction.} 
Collecting human annotations for structured tasks is costly, especially when discourse-level understanding and domain expertise are required~\cite{mieskes-stiegelmayr-2018-preparing,schulz-etal-2019-analysis,poudyal-etal-2020-echr}. It is thus desirable to reuse existing labels from a similar task, and transfer learning (TL) is often employed.
It can be divided into two broad categories~\cite{pan2009survey}: (1) {\em Transductive} approaches adapt models learned from a labeled source domain to a different target domain over the same task, and have shown promising results for discourse~\cite{kishimoto-etal-2020-adapting} and argument~\cite{chakrabarty-etal-2019-ampersand,accuosto-saggion-2019-transferring} related tasks.
(2) {\em Inductive} methods aim to leverage unlabeled data, usually in the same domain as the target domain, and have gained popularity with the pre-training and fine-tuning paradigm using Transformer models~\cite{devlin-etal-2019-bert,gururangan-etal-2020-dont}. 
We study both types in this work, with a particular focus on transductive approaches where the effect of different source domains are compared.

\paragraph{Active Learning (AL)} has been explored in many NLP problems including named entity recognition~\cite{tomanek2009reducing,shen2018deep}, text classification~\cite{tong2001support,hoi2006large}, and semantic parsing~\cite{iyer-etal-2017-learning,duong-etal-2018-active}.
Unlike the traditional supervised setting where training data is sampled beforehand, AL allows the learning system to actively select samples to maximize the performance, subject to an annotation budget~\cite{settles2009active,aggarwal2014active}.
Common AL strategies are either based on model uncertainty~\cite{houlsby2011bayesian,yuan-etal-2020-cold}, or promoting the diversity in sample distribution ~\cite{pmlr-v16-bodo11a,sener2018active}.
However, both paradigms require coupling sampled data with a specific learned model, which may cause subpar performance by a successor model~\cite{lowell-etal-2019-practical}. We propose model-independent acquisition strategies that are faster to train and do not rely on any model.

%% file: 03model.tex
\begin{figure}[t]
\centering
\includegraphics[width=75mm]{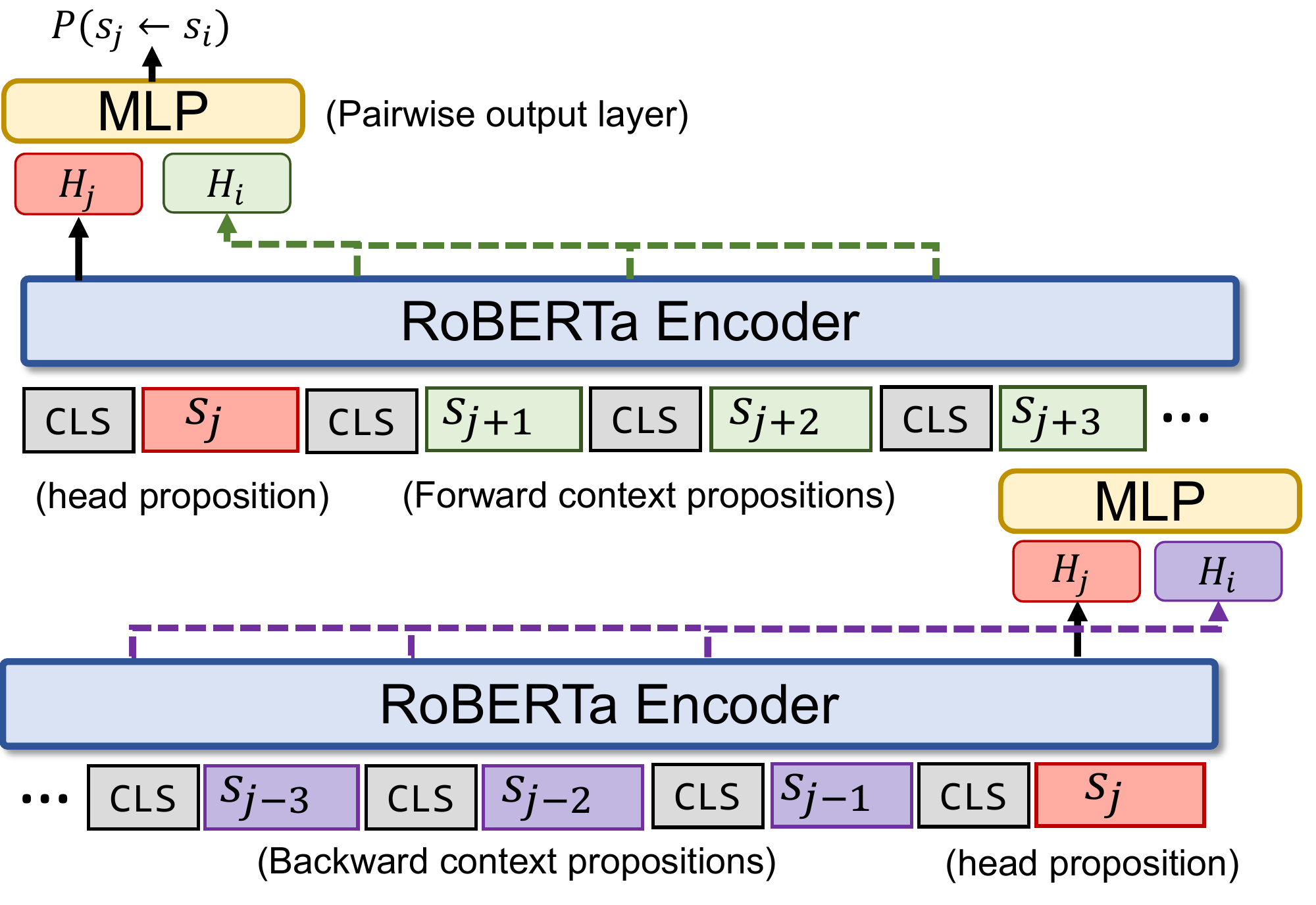}
\caption{
Our context-aware argument relation prediction model. 
For each head proposition $\bm{s}_j$, we encode both the backward (purple) and forward (green) contexts. 
$H_j$, the last layer states, represents proposition $\bm{s}_j$. 
$H_i$, where $i$ can be $j\pm1, j\pm2, \ldots, j\pm L$ ($L$ is the window size), is concatenated with $H_{j}$ and fed into the pairwise output layer, to yield the probability of $\bm{s}_{j} \leftarrow \bm{s}_i$. 
}
\label{fig:model}
\end{figure}

\paragraph{Task Formulation.} 
Given a document that is segmented into a list of propositions,
our task is to predict the existence of a \texttt{support} or \texttt{attack} link $\bm{s}_j \leftarrow \bm{s}_i$ between propositions $\bm{s}_i$ and $\bm{s}_j$. Here the targeted proposition $\bm{s}_j$ is the \textit{head}, and $\bm{s}_i$ is the \textit{tail}. 
{Our end-to-end model considers all proposition pairs.}
We also consider a simplified setting, where head propositions are given a priori.

\paragraph{A Context-aware Model.} 
Fig.~\ref{fig:model} depicts our model:
It is built on top of the RoBERTa encoder~\cite{liu2019roberta} which reads in a sequence of tokens. It contains stacked layers with bidirectional multi-headed self-attentions.
Different from prior work that only encodes single propositions, given a head proposition $\bm{s}_j$, we concatenate it with its surrounding context, including the $L$ propositions before and after it. Propositions are separated by \texttt{[CLS]} tokens. We use their last layer's states, denoted as ${H}_j$, to represent $\bm{s}_j$.
Other propositions within the window defined by $L$ then become candidates for tail propositions. 

After encoding, each tail candidate representation $H_i$ is concatenated with the head representation $H_j$ to form the input to the output layer, with the final prediction formulated as: 

{
\fontsize{9}{11}\selectfont
\setlength{\abovedisplayskip}{0pt}
\begin{align}
    P(y_r|\bm{s}_j, \bm{s}_i) = \text{softmax}(\text{tanh}{([H_j; H_i]\cdot\bm{W}_{1})}\cdot\bm{W}_{2}) \label{eq:output_layer}
\end{align}
}
where ${y_r}$ corresponds to three classes: \texttt{support}, \texttt{attack}, and \texttt{no-rel} if there is no link. $\bm{W}_{1}$ and $\bm{W}_{2}$ are trainable parameters.  
Dropout~\cite{JMLR:v15:srivastava14a} is added between layers. 

\smallskip
\noindent \textbf{Training objective} is cross-entropy loss over the labels of pairwise propositions within the context window. 
Our simplified setting reduces the prediction complexity from $\mathcal{O}(n^2)$~\cite{mayer2020ecai} to $\mathcal{O}(nL)$, with $n$ being the proposition count. 

%% file: 04active.tex
One major goal of this work is to explore AL solutions that can reduce the amount of samples for annotation, since labeling such a dataset can be the most laborious part of argument structure understanding. 
We consider a pool-based AL scenario~\cite{settles2009active}, where labels for the training set $\mathcal{U}$ are assumed to be unavailable initially. 
The learning procedure is carried out in $T$ iterations. In the $t$-th iteration, $b$ samples are selected using a given acquisition strategy. 
These samples are labeled and added into the labeled pool to comprise $\mathcal{D}_t$, on which a model $\mathcal{M}_t$ is then trained.

\subsection{Comparison Methods} 
For baselines, we consider \textbf{\textsc{random-prop}}, which samples $b$ propositions from the unlabeled training set with uniform distribution. Its variant, \textbf{\textsc{random}-}\textbf{\textsc{ctx}}, instead samples at the context level --- i.e., for a given head, its entire forward or backward context of $L$ propositions are sampled as a whole, until the total number of propositions reaches $b$.

The \textbf{\textsc{max-entropy}}~\cite{lewis1994sequential,joshi2009multi} method selects the most uncertain samples, based on the entropy score $\mathcal{H}(\cdot)$ using the model trained in the previous iteration:

{
\fontsize{9}{11}\selectfont
\setlength{\abovedisplayskip}{0pt}
\begin{align}
\mathcal{H}(y_r|\bm{s}_j,\bm{s}_i) = - \sum_r P(y_r|\bm{s}_j,\bm{s}_i) \text{log} P(y_r|\bm{s}_j,\bm{s}_i)
\end{align}
}
where $P(y_r|\bm{s}_j,\bm{s}_i)$ is the predicted probability of a relation label (Eq.~\ref{eq:output_layer}).

Bayesian Active Learning by Disagreement (\textbf{\textsc{bald}})~\cite{houlsby2011bayesian} is another common approach to exploit the uncertainty of unlabeled data by applying dropout at test time for multiple runs over the same sample, and picks ones with higher disagreement:

{
\fontsize{9}{11}\selectfont
\setlength{\abovedisplayskip}{0pt}
\begin{align}
\argmax_{\bm{s}_i}\mathcal{H}(y_r|\bm{s}_j, \bm{s}_i) - \mathbb{E}_{\theta}[ \mathcal{H}(y_r|\bm{s}_j, \bm{s}_i,\theta)]
\end{align}
\vspace{-3mm}
}

Uncertainty-based methods are at risk of selecting ``outliers'' or alike samples~\cite{settles2009active}. 
To encourage diversity of the selected samples, we consider \textbf{\textsc{coreset}}~\cite{sener2018active}, which enlarges differences among samples and achieves competitive performance in many vision tasks. 
At a high level, each sample is represented as a vector, e.g., we use the proposition representation $H_i$. 
A random set of $b$ samples are selected for labeling in the first iteration. 
In each subsequent iteration $t$, data points in the labeled pool $\mathcal{D}_{t-1}$ are treated as cluster centers, and the sample with the greatest $L_2$ distance from its nearest cluster center is selected. This process is repeated $b$ times to build the new labeled pool $\mathcal{D}_t$.

\subsection{Model-independent Acquisition Methods}
\label{sec:non-param-sampling}
One risk in AL is that samples selected by a model might not be useful for future models~\cite{lowell-etal-2019-practical}.
This motivates our design of \textit{model-independent} acquisition methods. 
Our first method, \textbf{\textsc{novel-vocab}}, promotes propositions with more unseen words. 
Assuming the frequency of a word $w$ in the labeled pool is $\mathcal{V}(w)$, the novelty score for an unlabeled sample $\bm{s}_i$ is computed as: 

{
\fontsize{9}{11}\selectfont
\setlength{\abovedisplayskip}{0pt}
\begin{align}
\text{novelty-score}(\bm{s}_i) = \sum_{w_t \in \bm{s}_i}\frac{f_{i,t}}{(1 + \mathcal{V}(w_t))}
\end{align}
}
where $f_{i,t}$ is the frequency of word $w_t$ in sample $\bm{s}_i$. Samples with the highest novelty scores are selected for labeling.  If a proposition has a high word overlap with samples in the labeled pool, the denominator $\mathcal{V}(w_t)$ will be high, and this sample is less likely to be chosen.

Our second method, \textbf{\textsc{disc-marker}}, aims to select more relation links by matching any of the following 18 prominent discourse markers from PDTB manual~\cite{webber2019penn} (matching statistics are in Appendix~\ref{sec:appendix-disc}).\footnote{When matched sentences exceed selection budget, we randomly sample with equal probabilities.}
For comparison, we also show a complementary approach \textbf{\textsc{no-disc-marker}}, which samples propositions \textit{without} any of those discourse markers. 

\begin{table}[h]
\centering

\begin{tabular}{|l|l|l|}
\hline
because & therefore & however\\
although & though & nevertheless \\
nonetheless & thus & hence \\
consequently & for this reason & due to \\
in particular & particularly & specifically \\
in fact & actually & but \\
\hline
\end{tabular}
\end{table}

%% file: 05data.tex
\begin{table}[t]
\fontsize{9}{11}\selectfont
\centering
\setlength{\tabcolsep}{0.6mm}
\begin{tabular}{l c c c c c}
     &  {\bf AMPERE++} & {\bf Essays} & {\bf AbstRCT} & {\bf ECHR} & {\bf CDCP} \\
    \midrule
    \# Doc. & 400 & 402 & 700 & 42 & 731 \\
    \# Tok. & 190k & 147k & 236k & 177k & 89k \\
    \# Prop. & 10,386 & 12,373 & 5,693 & 6,331 & 4,932 \\
    \# Supp. & 3,370 & 3,613 & 2,402 & 1,946 & 1,426 \\
    \# Att. & 266 & 219 & 70 & 0 & 0 \\
    \# Head & 2,268 & 1,707 & 1,138 & 741 & 1,037 \\
    Density & 21.8\% & 13.8\% & 20.0\% & 11.7\% & 21.0\% \\
    \bottomrule
\end{tabular}

\caption{
Statistics of five datasets, including our AMPERE++ data with newly annotated relations on AMPERE~\cite{hua-etal-2019-argument}. We report the total numbers of documents (\# Doc.), tokens (\# Tok.), propositions (\# Prop.), support (\# Supp.) and attack (\# Att.) relations, unique head propositions (\# Head), and relation density as the percentage of propositions that are supported or attacked by at least one proposition. 
}
\label{tab:stats}
\end{table}

\begin{figure*}[t]
\centering
\includegraphics[width=155mm]{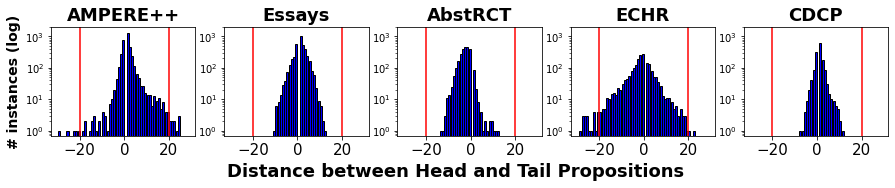}
\caption{
Distribution of distance (measured by number of propositions) between head-tail pairs across five domains.  Positive values indicate that the tail appears after the head in the document, and vice versa.
}
\label{fig:distance}
\end{figure*}

We experiment with five datasets from distinct domains, with key statistics listed in Table~\ref{tab:stats}. Below we outline data collection and annotation, notable preprocessing steps, and data splits.

\paragraph{Domain 1: Peer Reviews (New Annotation).}
We first annotate argument relations on AMPERE~\cite{hua-etal-2019-argument}, which consists of $400$ ICLR 2018 paper reviews collected from OpenReview. 
Each review has been annotated with segmented propositions and corresponding types (i.e., \textit{evaluation}, \textit{request}, \textit{fact}, \textit{reference}, and \textit{quote}). 
We augment this dataset by labeling the support and attack relations among the propositions. This new dataset is called \textbf{\textsc{AMPERE++}}. 

We hire three proficient English speakers to annotate the entire dataset in multiple rounds. 
During annotation, they are displayed with the propositions along with their types. We impose two constraints. 
(1) Each proposition can only support or attack at most one other proposition. 
(2) Factual propositions ({\it fact}, {\it reference}, {\it quote}) cannot be supported or attacked by subjective ones ({\it evaluation}, {\it request}). Similar rules are used by \newcite{PARK18.679}. 
We include detailed guidelines in Appendix~\ref{sec:appendix-annotation}. 
For quality control and disagreement resolution, the annotators are joined by a fourth judge after each round, where they discuss samples with different labels to reach agreement.

The resulting dataset contains 3,636 relations from 400 reviews with a substantial inter-annotator agreement score of 0.654 (Fleiss' $\kappa$).
Following our prior work~\cite{hua-etal-2019-argument}, we use 300 reviews for training, 20 for validation, and 80 for test. 
We also collect 42k reviews from OpenReview for ICLR 2019-2021, UAI 2018, and NeurIPS 2013-2020, which are used in the self-supervised learning experiments for improving representation learning.

\paragraph{Domain 2: Essays.}
Our second dataset is based on the essays curated by \newcite{stab-gurevych-2017-parsing} from \url{essaysforum.com}. Argumentative propositions are identified at the sub-sentence level and labeled as ``{\it premise}'', ``{\it claim}'', or ``{\it major claim}''. 
Support and attack relations are annotated from a premise to a claim or to another premise. The link cannot cross paragraph boundaries, highlighting the dataset's focus on relations close by.  

We split the original training set into 282 essays for training and 40 for validation. The remaining 80 are reserved for test. Similarly, we also download 26K essays from the same online forum for self-supervised representation learning. 

\paragraph{Domain 3: Biomedical Paper Abstracts.}
Next, we use the \textbf{AbstRCT} corpus~\cite{mayer2020ecai}, which contains 700 paper abstracts retrieved from PubMed.\footnote{\url{https://pubmed.ncbi.nlm.nih.gov/}} 
The primary subjects are Randomized Controlled Trials of diseases.
Notably, AbstRCT has much fewer propositions and relations than the previous two datasets, due to the factual nature of paper abstracts. 

Following \newcite{mayer2020ecai}, we use 350 abstracts for training, 50 for validation, and 300 for test. 
We employ the 133K unlabeled abstracts released by \newcite{cohan-etal-2018-discourse} for self-supervision.

\paragraph{Domain 4: Legal Documents.}
Legal texts are studied in the early work of argument mining~\cite{palau2009argumentation,lippi2016argumentation}. We choose the {\bf ECHR} corpus~\cite{poudyal-etal-2020-echr}, containing 42 recently-annotated case-law documents of the European Court of Human Rights.
The authors define an argument structure as a list of premises and a conclusion. 
We consider each premise as linked to the corresponding conclusion. 
The dataset is split into 27 documents for training, 7 for validation, and 8 for test.

\paragraph{Domain 5: Online User Comments.}
Finally, we include the Cornell eRulemaking Corpus~\cite{PARK18.679}, extracted from an online forum where the public argues for or against proposed rules.
The 731 annotated comments are mostly related to the Consumer Debt Collection Practices rule ({\bf CDCP}), and is annotated with support relations only. 
We adopt the original splits: 501 for training, 80 for validation, and 150 for test.
On average, there are less than two relation links per comment, and  only 21\% of the propositions are supported.

\paragraph{Head-tail Distance Distribution.}
\label{sec:data-distance}
Recall that our context-aware model only encodes context propositions up to a fixed window size.
Although this setup neglects some relation links, we show in Fig.~\ref{fig:distance} that a large enough window size (e.g., 20) is sufficient to cover all (Essays, CDCP, AbstRCT) or over 98\% (AMPERE++, ECHR) of all relations. 

Fig.~\ref{fig:distance} further highlights domain-specific patterns.
AMPERE++ and CDCP are skewed to the right, indicating reviewers and online users tend to put their claims upfront with supporting arguments appearing later. 
On the contrary, paper abstracts (AbstRCT) usually describe premises first and then draw conclusions. 
Essays and ECHR have more balanced distributions between both directions.

\paragraph{Proposition Length and Label Distribution.}
Due to differences in argument schemes, proposition length varies considerably across domains. 
AbstRCT has the longest propositions with an average of 45 tokens. 
Consequently, the actual encoder input may contain less than 20 propositions due to the maximum token limit.
Under our context-aware encoding, the ratio of positive samples (\texttt{support} or \texttt{attack}) is boosted to 29\% because they are less likely to be truncated due to the relative proximity to head propositions (Fig.~\ref{fig:distance}). The other four domains have similar positive ratios, ranging from 6\% (AMPERE++) to 17\% (CDCP).

Existing relation prediction methods~\cite{stab-gurevych-2017-parsing,niculae-etal-2017-argument,mayer2020ecai} label all pairwise propositions within the same document, leading to much lower positive ratios, especially for ECHR where documents are long. In \S\ref{sec:results-supervised} we show that such unbalanced distribution poses difficulties for traditional methods.

%% file: 06experiment.tex
In this section, we design experiments to answer the following questions.  
(1) To which degree is the context-aware model better at identifying argumentative relations (\S\ref{sec:results-supervised})? 
(2) How much improvement can transfer learning (TL) make when different source domains are considered for a target domain (\S\ref{sec:results-tl})?
(3) Does unlabeled in-domain data help downstream tasks using self-supervised pre-training and inductive transfer learning (\S\ref{sec:results-tl})?
(4) How do active learning (AL) strategies perform on relation prediction and whether combining transfer learning leads to further performance boost (\S\ref{sec:results-al})?

\smallskip
\noindent \textbf{Evaluation} is based on macro-F1 scores as done in prior work~\cite{stab-gurevych-2017-parsing,niculae-etal-2017-argument}. For tasks without attack labels (ECHR and CDCP), the macro average is calculated over \texttt{support} and \texttt{no-rel} only, otherwise it is averaged over three classes.
Each setup is run five times with different random seeds, and the average scores on test sets are reported.

\smallskip
\noindent \textbf{Implementation} of our models is based on the Transformer~\cite{wolf-etal-2020-transformers}. 
Our encoder is RoBERTa-base~\cite{liu2019roberta}, which has 12 layers with a hidden size of 768.
We apply dropout~\cite{JMLR:v15:srivastava14a} with a probability of 0.1 for the output MLP layer. 
We use the Adam optimizer~\cite{kingma:adam} with 16 sequences per batch. 
We hyper-tune our proposed argument relation prediction model with different number of maximum training epochs $\{\text{5}, \text{10}, \text{15}\}$, warmup steps $\{\text{0}, \text{1000}, \text{5000}\}$, learning rate $\{\text{1e-5}, \text{1e-6}, \text{5e-5}\}$, and scheduler $\{\texttt{constant}, \texttt{linear}\}$. The best validation result is achieved with 15 epochs, 5000 warmup steps, 1e-5 as learning rate, and the \texttt{constant} scheduler. We use this configuration for all model training experiments.

\subsection{Supervised Learning Results}
\label{sec:results-supervised}

We first evaluate our model with the standard supervised learning over the full training set using varying window sizes. 
We assume the heads are given at both training and inference, except for the end-to-end setting.

\paragraph{Comparisons.}
We implement an \textbf{\textsc{SVM}} with features adapted from Table 10 of \newcite{stab-gurevych-2017-parsing},
except for features specific to the essays domain (e.g., whether a proposition is in the introduction). 
We experiment with both linear and radial-basis function (RBF) kernels, with regularization coefficients tuned on validation. 
More details can be found in Appendix~\ref{sec:appendix-svm}. 

\textbf{\textsc{SeqPair}} is
based on the sequence pair classification setup~\cite{devlin-etal-2019-bert} using the pre-trained RoBERTa. 
Each pair of head and tail is concatenated and segmented with the \texttt{[SEP]} token. 
The \texttt{[CLS]} token is prepended to the beginning of the sequence and used for classification. This setup resembles the model in \newcite{mayer2020ecai}.

We further compare with two dataset-specific \textbf{\textsc{benchmark}} models:  \newcite{stab-gurevych-2017-parsing} use a rich set of features tailored for essays to train SVMs, and \newcite{niculae-etal-2017-argument} employ structured SVMs on CDCP.

\begin{table}[t]
\centering
\fontsize{9}{11}\selectfont
\setlength{\tabcolsep}{0.2mm}
\begin{tabular}{l c c c c c c}

\toprule
    & {\bf AMPERE++} & {\bf Essays} & {\bf AbstRCT} & {\bf ECHR} & {\bf CDCP}   \\
    \midrule
    \textsc{SVM}-linear  & 24.82 & 28.69 & 33.60  & 21.18 & 29.01 \\
    \textsc{SVM}-RBF & 26.38 & 31.68  & 32.65 & 21.36 & 30.34 \\
    \textsc{SeqPair} & 23.40 & 38.37 & {\bf 66.96} & 13.76 & 35.23  \\
    \textsc{Benchmark} & - & {\bf 73.30} & - & - & 26.70  \\
    \hline
    \multicolumn{6}{l}{\textsc{Ours} (head given)} \\
    \quad $L = 5$ & 66.34 & 65.61 & 55.48 & 60.92 & 64.82 \\
    \quad $L = 10$ & 75.69 & 69.41 & 59.27 & 67.51 & 69.47 \\
    \quad $L = 20$ & {\bf 77.64} & 71.30 & 63.62 & {\bf 70.82} & {\bf 70.37} \\
    \hdashline
    \multicolumn{6}{l}{\textsc{Ours} (end-to-end)} \\
    \quad $L = 20$  & 74.34 & 67.68 & 63.73 & 61.35 & 63.13 \\
    \bottomrule
\end{tabular}

\caption{
F1 scores for argument relation prediction. 
Each entry is averaged over five runs with different random seeds.
The best result for each dataset is {\bf bolded}. {Our context-aware model outperforms both baselines except for AbstRCT. The difference between {\em head given} and {\em end-to-end} is close, suggesting that the key challenge for structure extraction lies in relation prediction. Our model performance improves when larger window size $L$ is used.} 
}
\label{tab:supervised-results}
\end{table}

\paragraph{Results.}
As shown in Table~\ref{tab:supervised-results}, our context-aware model outperforms the comparisons except for Essays and AbstRCT.
The feature-rich SVM marginally outperforms our model, though the features are not generalizable to new domains.
As mentioned in \S\ref{sec:data}, AbstRCT has much higher positive ratio than other domains. This indicates that our model is more robust against unbalanced training data than the pairwise approach.

The performance drop for end-to-end models are marginal in most cases, underscoring relation prediction as the key challenge for structure extraction, which the simplified setup has to tackle as well.

\subsection{Transfer Learning Results}
\label{sec:results-tl}

Results in the previous section show large performance discrepancies among different domains. For instance, domains with few labeled samples, such as AbstRCT and CDCP, lead to worse performance. 
Moreover, annotating argument structures for some domains is even more involved, e.g., \newcite{poudyal-etal-2020-echr} hired three lawyers to annotate ECHR legal documents. 
We hypothesize that basic reasoning skills for understanding argument structures can be shared across domains,
thus we study transfer learning, a well-suited technique that leverages existing data with similar task labels ({\em transductive}) or unlabeled data of the same target domain ({\em inductive}).
Concretely, we present thorough experiments of TL over all transfer pairs, where the model is first trained on the source domain and fine-tuned on the target domain.

\paragraph{Transductive TL.}
The upper half of Table~\ref{tab:tl_results} shows that \textit{three out of four models transferred from AMPERE++ achieve better performance than their supervised learning counterparts in Table~\ref{tab:supervised-results}}. 
In particular, we observe more than 5 F1 points gains on ECHR and CDCP, which contain the least amount of labeled samples. 
However, when transferred from the four other datasets, performance occasionally drops. 
This can be due to the distinct language style and argumentative structure (AbstRCT), the source domain size (CDCP, ECHR), or the model's failure to learn good representations due to over-reliance on discourse markers (Essays). 
Overall, \textit{AMPERE++ consistently benefits diverse domains for argument structure understanding, demonstrating its usage for future research}.  

\definecolor{hl_cell_color}{RGB}{221,251,221}
\begin{table}[t]
\centering
\setlength{\tabcolsep}{0.2mm}
\fontsize{9}{11}\selectfont
\begin{tabular}{l c c c c c}
\toprule
     & {\bf AMPERE++} & {\bf Essays} & {\bf AbstRCT} & {\bf ECHR} & {\bf CDCP}   \\
    \rowcolor{gray!20} 
    \multicolumn{6}{l}{SRC $\rightarrow$ TGT (Transductive TL)} \\
    
    {\bf AMPERE++} & -- & \hlc[hl_cell_color]{73.84} & {63.42} & \hlc[hl_cell_color]{\bf 76.50}  & \hlc[hl_cell_color]{\bf 75.93}   \\
    {\bf Essays}  & \hlc[hl_cell_color]{77.93} & -- & 60.62  & 68.72 & \hlc[hl_cell_color]{74.11}\\
    {\bf AbstRCT} & 76.29 & 71.17 & -- & \hlc[hl_cell_color]{73.31} & 69.17 \\
    {\bf ECHR} & \hlc[hl_cell_color]{77.69} & 70.82 & 47.91 & -- & 69.30 \\
    {\bf CDCP} & \hlc[hl_cell_color]{77.87} & 68.37 & 62.38 & \hlc[hl_cell_color]{72.03} & --  \\
      \rowcolor{gray!20} 
     \hline\hline
    \multicolumn{6}{l}{TGT-pret $\rightarrow$ TGT (Inductive TL)} \\
    {\bf MLM} & \hlc[hl_cell_color]{78.10} & \hlc[hl_cell_color]{74.21} & \hlc[hl_cell_color]{\bf 64.48} & -- & -- \\
    {\bf Context-Pert} & \hlc[hl_cell_color]{79.01} & 68.36 & 59.47 & -- & -- \\
    \hline\hline
    \rowcolor{gray!20} 
    \multicolumn{6}{l}{SRC-pret $\rightarrow$ SRC $\rightarrow$ TGT} \\
    {\bf AMPERE++} & -- & 70.42 & 61.84 & \hlc[hl_cell_color]{70.96} & \hlc[hl_cell_color]{74.82}   \\
    {\bf Essays} & 44.40 & -- & 58.59 & \hlc[hl_cell_color]{73.58}  & \hlc[hl_cell_color]{71.84}  \\
    {\bf AbstRCT} & 76.25 & 69.26 & --  & \hlc[hl_cell_color]{70.93}  & \hlc[hl_cell_color]{71.67}   \\
    
    \rowcolor{gray!20} 
    \multicolumn{6}{l}{TGT-pret $\rightarrow$ SRC $\rightarrow$ TGT} \\
    {\bf AMPERE++} & -- & \hlc[hl_cell_color]{\bf 74.90} & 62.34 & --  & --  \\
    {\bf Essays} & 76.69 & -- & 62.38 & --  & --  \\
    {\bf AbstRCT} & \hlc[hl_cell_color]{\bf 79.52} & \hlc[hl_cell_color]{73.09}  & -- & --  & --  \\

    \bottomrule
\end{tabular}
\caption{
Results for transfer learning. First column denotes the source domain, the rest are target domains. 
The best result per column is in {\bf bold}. 
Transfer learning that outperforms the in-domain training setup (Table~\ref{tab:supervised-results}, second last row) is highlighted in \hlc[hl_cell_color]{green}. Notably, using AMPERE++ as the source domain yields better performance than the standard supervised setting. Overall, self-supervised pre-training can further benefit transductive transfer learning.
} 
\label{tab:tl_results}
\end{table}

\begin{figure}
    \centering
    \includegraphics[width=75mm]{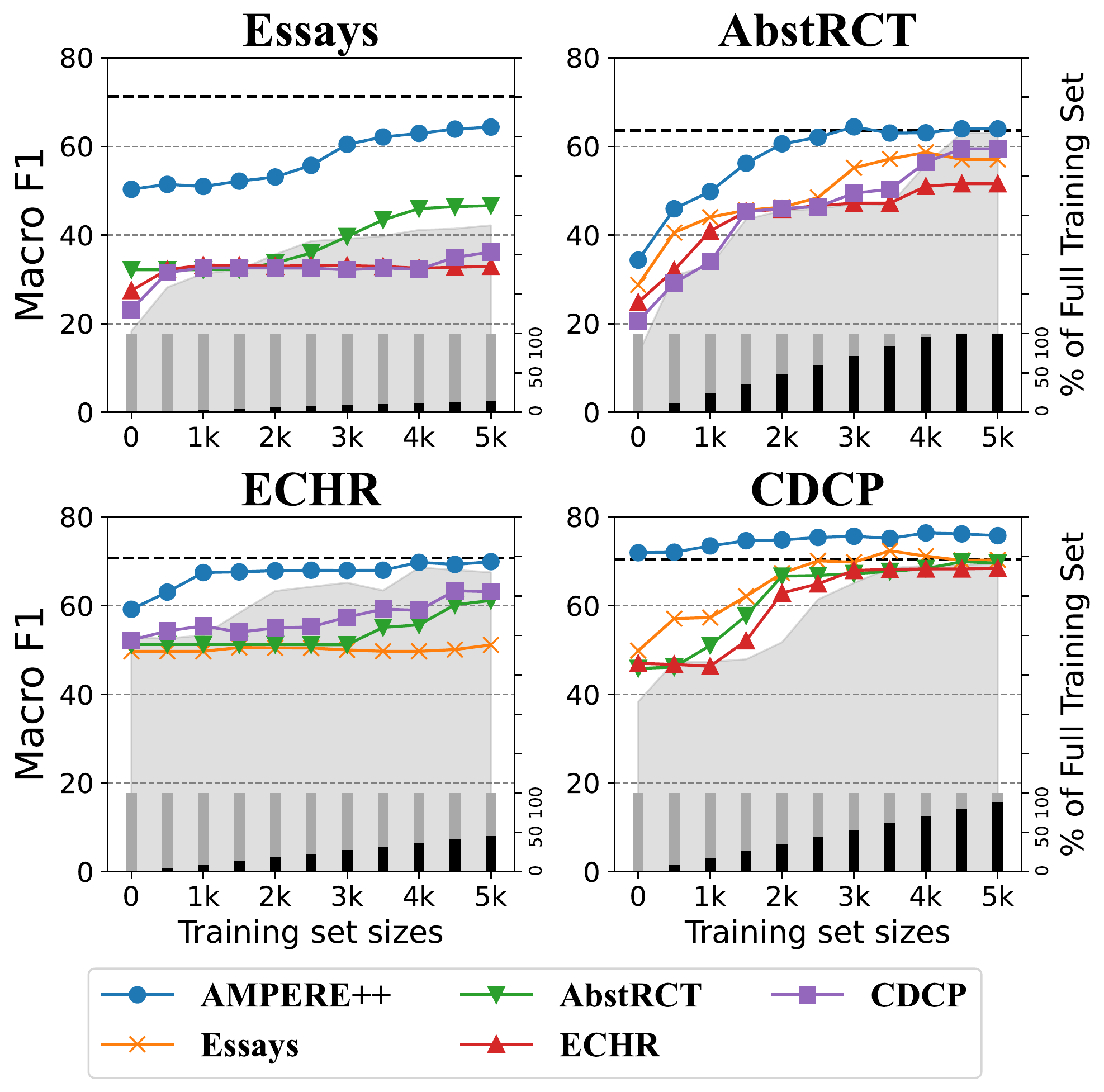}
    \caption{
    Macro F1 scores with limited training data.
    We sample training set from 0 to 5,000 samples, in an increment of 500.
    Bottom bars indicate the percentage of such subsets over the full training set.
    Scatter plots represent the transfer learning results from different source domains, with those from non-TL settings marked as shaded areas. 
    Horizontal dashed lines represent the performance using the full training set.
    Models using AMPERE++ as the source domain consistently yield better F1 scores than others and non-TL models.
    }
    \label{fig:tl_overall}
\end{figure}

\paragraph{Inductive TL.}
Motivated by recent findings~\cite{beltagy-etal-2019-scibert,lee2020biobert,gururangan-etal-2020-dont} that self-supervised pre-training over specific domains significantly improves downstream tasks, we also consider the {\em inductive} transfer learning setup with the following two objectives:
(1) masked language model (\textbf{MLM}) prediction, which randomly selects $15\%$ of the input tokens for prediction as done in~\newcite{devlin-etal-2019-bert}; 
(2) \textbf{context-aware sentence perturbation} ({\bf Context-Pert}), which packs each document into a sequence of sentences segmented by the \texttt{[CLS]} token, $20\%$ of which are replaced by random sentences from other documents, another $20\%$ shuffled within the same document, and the rest unchanged. The pre-training objective is to predict the perturbation type of each sentence. 
Results are in the middle part of Table~\ref{tab:tl_results}, where MLM pre-training benefits all three domains. Context-Pert improves AMPERE++ even more, but negatively affects the other two domains.

\begin{figure*}
    \centering
    \includegraphics[width=160mm]{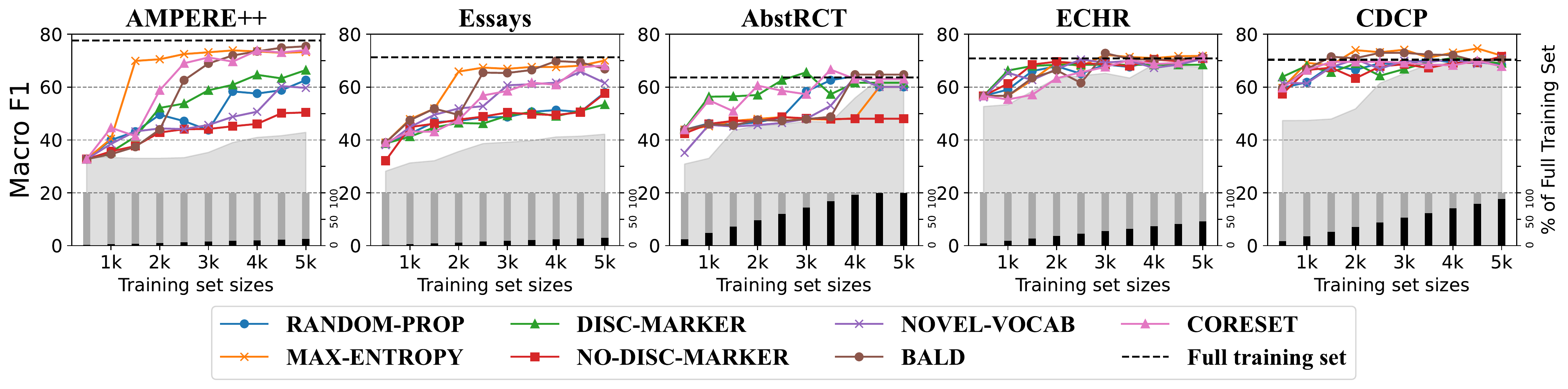}
    \caption{Active learning results using different acquisition methods in 10 iterations. Shaded areas stand for the \textbf{\textsc{random-ctx}} performance, which aligns with that in Figure~\ref{fig:tl_overall}.
    We show performance for three model-independent strategies, \textbf{\textsc{disc-marker}}, \textbf{\textsc{novel-vocab}}, \textbf{\textsc{no-disc-marker}}, alongside three strong comparisons.
    The model-independent strategies yields significantly better results than random sampling.
    On AbstRCT, ECHR, and CDCP, \textbf{\textsc{disc-marker}} achieves better or competitive performance than \textbf{\textsc{max-entropy}} and \textbf{\textsc{bald}}. 
    To better visualize the performance difference, rescaled plots for ECHR and CDCP are in Appendix~\ref{sec:appendix-rescaled-al}.
    }
    \label{fig:al}
\end{figure*}

\paragraph{Combining Inductive and Transductive TL.}
Moreover, we showcase that \textit{adding self-supervised learning as an extra pre-training step for transductive TL further boosts performance.} 
From the lower half of Table~\ref{tab:tl_results}, the pre-trained model uniformly improves over the standard transductive TL. 
Notably, \textit{using target domain for pre-training leads to better results than using the source domain data}.
This implies that better representation learning for target domain language is more effective than a stronger source domain model.

\paragraph{Effectiveness of TL in Low-Resource Setting.}
To quantitatively demonstrate how TL benefits low-resource target domains, we control the size of training data and conduct transductive TL for each domain. Fig.~\ref{fig:tl_overall} plots the trends where training data varies from 0 to 5,000, incremented by 500.
{\em Among all datasets, AMPERE++ yields the best transfer learning results as the source domain: Using less than half of the target training set, it allows to approach or exceed the fully trained models.} 
For other datasets, we observe mixed results when they are used as the source. In general, TL brings more improvements when less training data is used.

\begin{figure}[t]
    \centering
    \includegraphics[width=75mm]{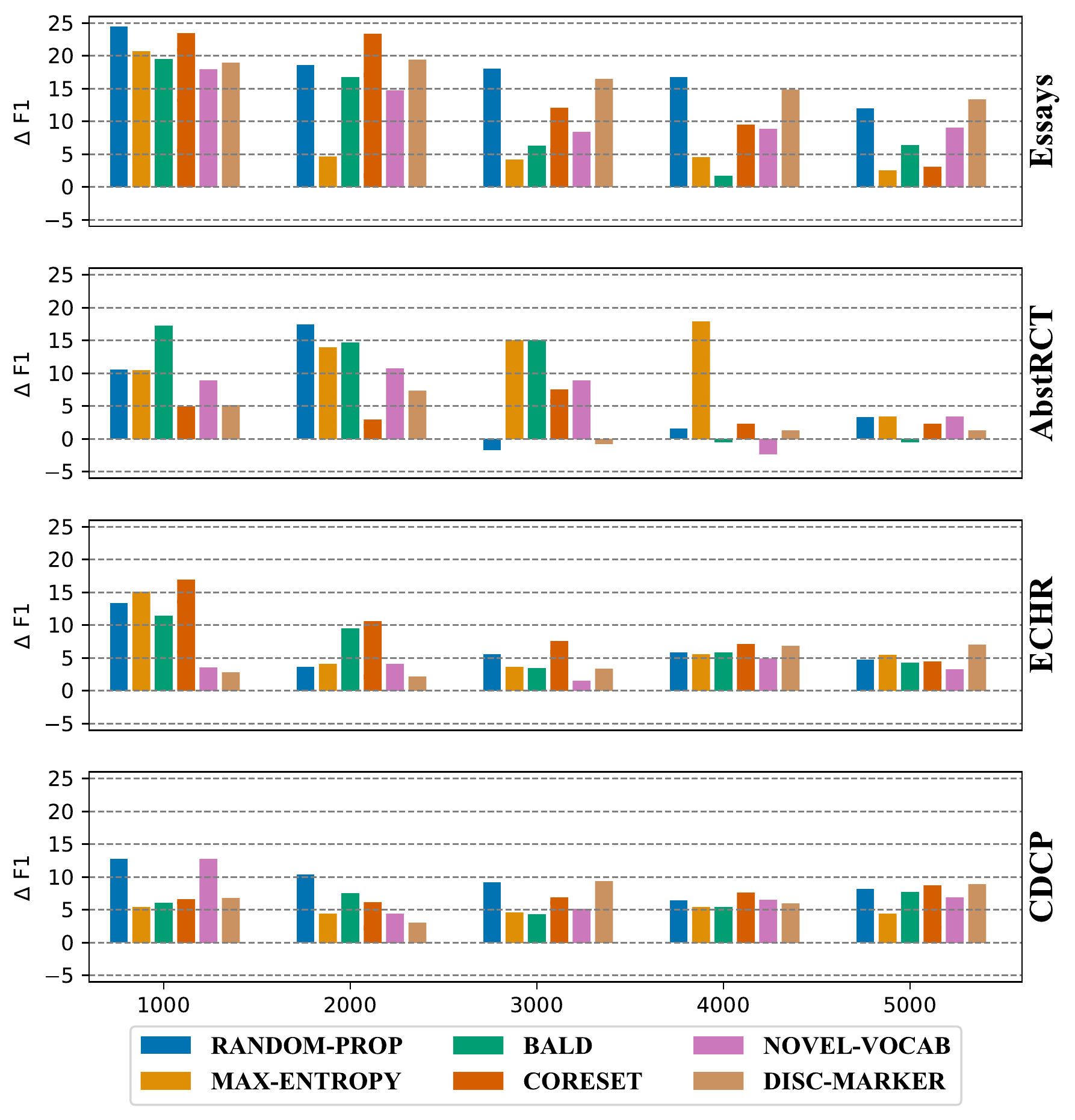}
    \caption{
    Improvements of macro F1 scores by adding TL to each AL strategy. AMPERE++ is used as the source domain for TL. We observe consistent gains across the board except for AbstRCT when the training samples are close to full. Generally, the improvements decline when more training samples are included.}
    \label{fig:al_tl_delta}
\end{figure}

\subsection{Active Learning Results} 
\label{sec:results-al}

\paragraph{Comparisons of Acquisition Strategies.} 
Fig.~\ref{fig:al} plots the F1 scores for all strategies as discussed in \S\ref{sec:al} across 10 AL iterations. 
As expected, the performance gradually improves with more labeled data.
The three model-based methods: \textbf{\textsc{max-entropy}}, \textbf{\textsc{bald}}, and \textbf{\textsc{coreset}} generally attain better performance, suggesting the efficacy of common AL methods on argument relation understanding.
The model-independent strategies yield competitive results. 
In particular, \textbf{\textsc{disc-marker}} proves to be a good selection heuristics for AMPERE++ and AbstRCT. Its relatively low scores on Essays is likely due to the abundance of discourse markers in this domain, so that random sampling would have similar effects.
By contrast, avoiding discourse markers (\textbf{\textsc{no-disc-marker}}) tends to hurt performance.
Notably, \textit{without relying on any trained model, task-specific acquisition strategies can be effective for labeling argument relations}.

\paragraph{Warm-start Active Learning.}
Finally, we investigate the added benefits of transfer learning for major active learning systems.
In each AL iteration, we warm-start the model with checkpoints trained from AMPERE++, and calculate the difference of F1 scores from the non-TL counterpart. Fig.~\ref{fig:al_tl_delta} shows the results for five of the ten iterations. 
We observe improvements across the board, especially with small training data size.
For AbstRCT, the TL warm-start either makes no difference or slightly hurts performance after 3,000 samples are available, whereas the \textbf{\textsc{max-entropy}} method constantly benefits from warm-starting. Our findings suggest that TL is an effective add-on for early stage AL, benefiting different strategies uniformly.

%% file: 07conclusion.tex
We present a simple yet effective framework for argument structure extraction, based on a context-aware Transformer model that outperforms strong comparisons on five distinct domains, including our newly annotated dataset on peer reviews. 
We further investigate two complementary frameworks based on transfer learning and active learning to tackle the data scarcity issue. 
Based on our extensive experiments, transfer learning from our newly annotated AMPERE++ dataset and self-supervised pre-training consistently yield better performance. Our model-independent strategies approach popular model-based active learning methods.

%% file: 08acknowledge.tex
This research is supported in part by National Science Foundation through Grant IIS-2100885. 
We thank four anonymous reviewers for their valuable suggestions on various aspects of this work.

%% file: 09appendix.tex
\section{Model and General Details}

\subsection{Discourse Markers}
\label{sec:appendix-disc}
In \S\ref{sec:non-param-sampling} of the main paper we introduce a \textbf{\textsc{disc-marker}} based acquisition method for active learning. 
The matching statistics of the 18 discourse markers are shown in Fig.~\ref{fig:disc-marker-matched}.
We break down the count based on whether a proposition is the \texttt{head} or \texttt{tail} of any relation. As expected, certain discourse markers such as ``because'', ``but'', and ``due to'' likely indicate a \texttt{tail} proposition, whereas ``therefore'', ``thus'' tend to be found in \texttt{head} propositions.

\subsection{SVM Comparison}
\label{sec:appendix-svm}

In Table~\ref{tab:svm-feat}, we describe the full feature set used in the SVM comparison model in \S~6.1 of the main paper. These features are adapted from Table 10 of \newcite{stab-gurevych-2014-identifying}. The indicators are from their Table B.1 in the Appendix.

For hyper-parameter search, we tune the regularization coefficient $C$ over values $\{0.1, 0.5, 1.0, 10.0\}$. The best performing model (macro-F1) on validation set is used for evaluation.

\begin{table}[h]
\centering
\fontsize{10}{11}\selectfont
\begin{tabular}{lp{55mm}}
\toprule

\textbf{Group} & \textbf{Description} \\
\midrule

Lexical & Binary lemmatized unigram of the head and tail propositions (top 500 frequent ones are considered)  \\
Syntactic & Binary POS features of head and tail propoisitions \\
Structural & Number of tokens of head and tail; Number of propositions between source and tail; head presents before tail; tail presents before head \\
Indicator & Indicator type present in head or tail; indicator type present between head and tail \\
ShNo & Shared nouns between head and tail propositions (number and binary) \\

\bottomrule

\end{tabular}
\caption{Features used for SVM model.}
\label{tab:svm-feat}
\end{table}

\subsection{Active Learning Results}
\label{sec:appendix-rescaled-al}

\begin{figure}[t]
    \centering
    \includegraphics[width=78mm]{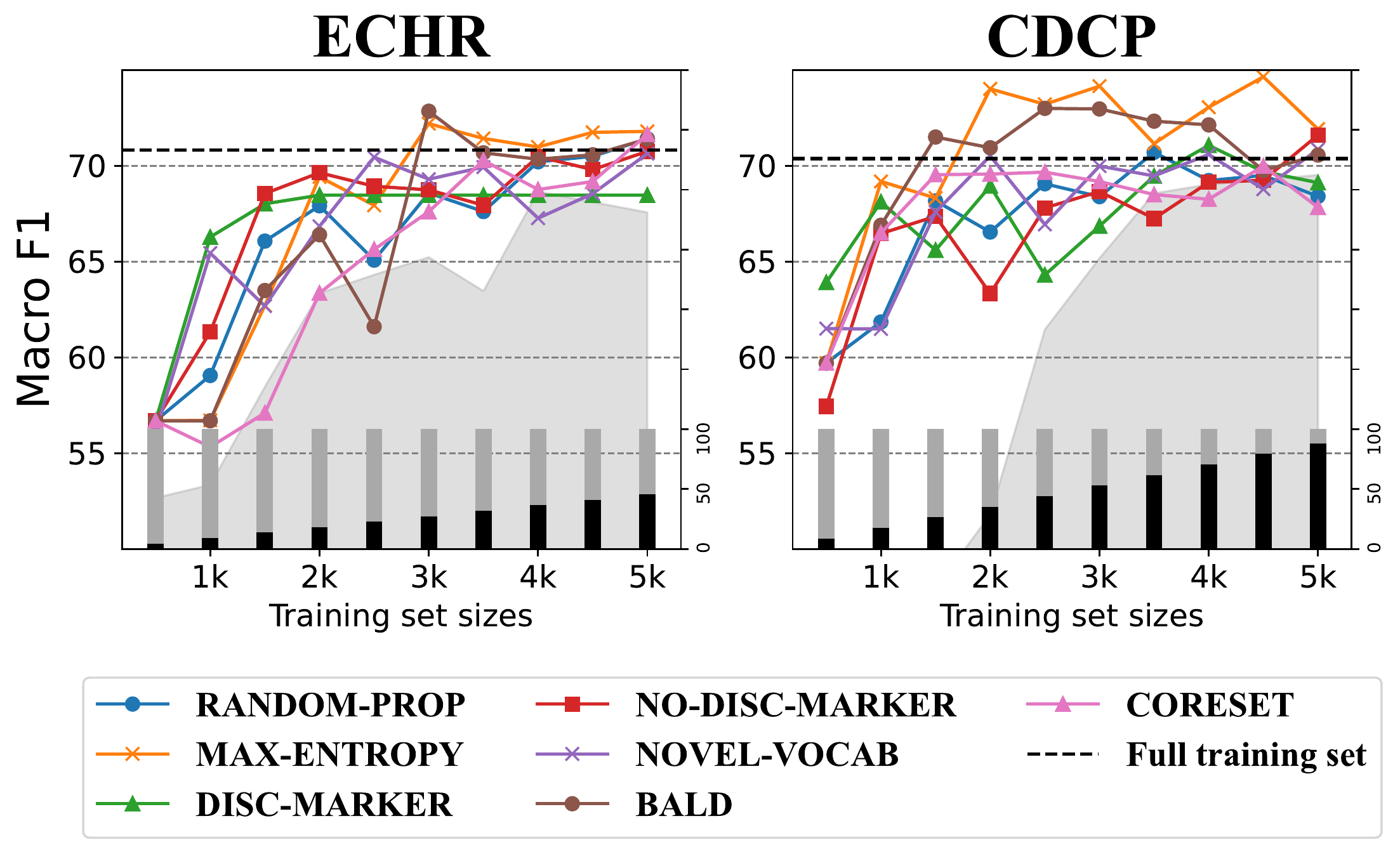}
    \caption{Active learning results for ECHR and CDCP on 50--80 scale. The scores are the same as the rightmost two plots in Figure~\ref{fig:al}.}
    \label{fig:al_rescaled}
\end{figure}

\begin{figure*}[t]
    \centering
    \includegraphics[width=155mm]{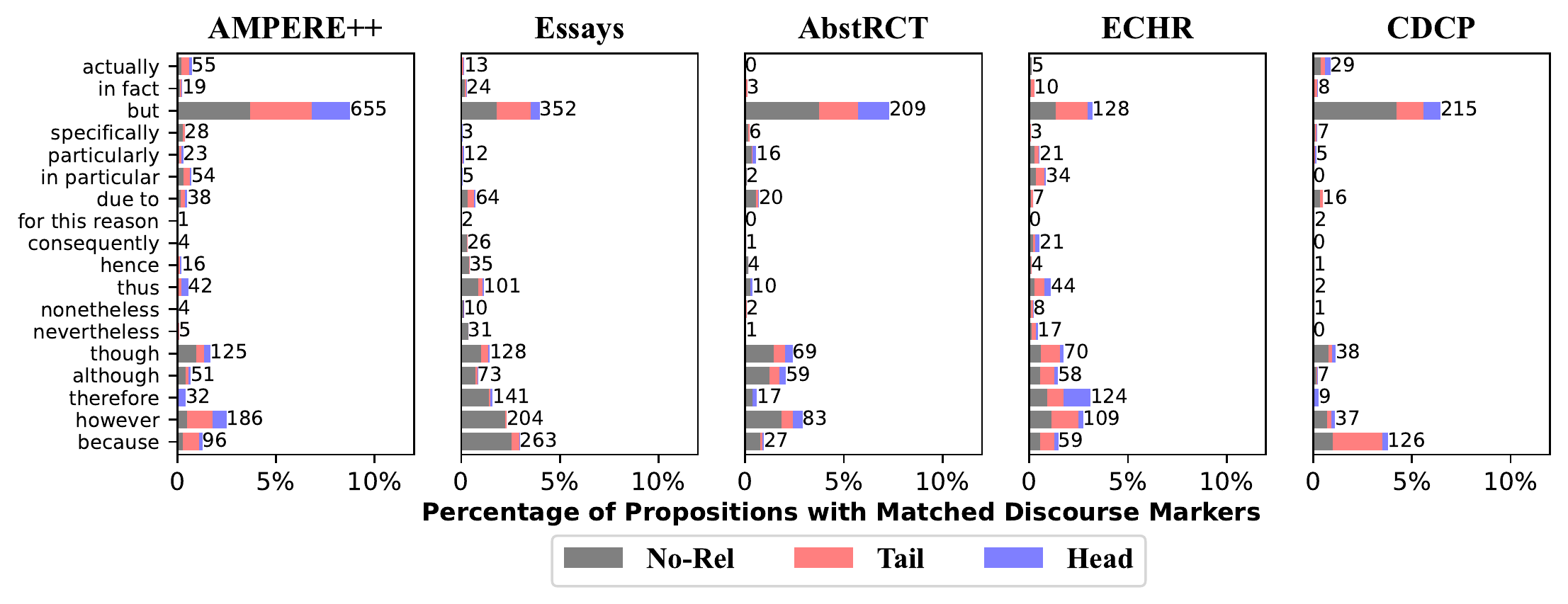}
    \caption{
    The distribution of matched discourse markers in each dataset. We indicate the raw count next to each bar. Propositions that are \texttt{tail} or \texttt{head} of any relation are highlighted in colors. Overall, about 10--20\% of the propositions contain at least one discourse marker. Certain discourse markers correlate well with the existence of argument relations. For instance, ``because'', ``due to'', ``however'' are more likely to be found in \texttt{tail}; ``therefore'', ``thus'' tend to appear in \texttt{head}.}
    \label{fig:disc-marker-matched}
\end{figure*}

In Fig.~\ref{fig:al}, we compare active learning methods over five datasets on the same 0-80 scale. Results of different strategies fall in tight ranges for ECHR and CDCP. For better visualization, we show the same figure on a 50--80 scale in Fig.~\ref{fig:al_rescaled}.

\section{AMPERE++ Annotation}
\label{sec:appendix-annotation}

To annotate argument relations over the AMPERE~\cite{hua-etal-2019-argument} dataset, we hire three proficient English speakers who are US-based college students. The first author serve as the judge to resolve disagreements.
The detailed guidelines are shown in Table~\ref{tab:anno}. Throughout the annotation, we identify difficult cases and summarize representative ones in Table~\ref{tab:appendix-difficult}.

\begin{table*}[t]
\fontsize{10}{11}\selectfont
\centering
\begin{tabular}{p{6mm}p{140mm}}
\toprule
    {\bf Tail} & Only macro-average F-scores are reported. \\
    {\bf Head} & Please present micro-average scores as well  \\
    {\bf Label} & \texttt{support} \\
    \midrule
    {\bf Tail} & Fig 3: This could really be drawn considerably better \\
    {\bf Head} & Make the dots bigger and their colors more distinct.  \\
    {\bf Label} & \texttt{support} \\
    \midrule
    {\bf Tail} & Fig 4. right looks like a reward signal. \\
    {\bf Head} & but is labelled Proportion correct.  \\
    {\bf Label} & \texttt{attack} \\
    \midrule
    {\bf Tail} & This idea is not novel \\
    {\bf Head} & In the first part of the paper (Section 2) the authors propose to use the optimal transport distance $\ldots$ as the objective for GAN optimization.  \\
    {\bf Label} & \texttt{attack} \\
    \midrule
    {\bf Tail} & Then, the difference is crystal clear. \\
    {\bf Head} & The difference between Figure 1, 4, and 6 could be clarified.  \\
    {\bf Label} & \texttt{no-rel} \\
    \midrule
    {\bf Tail} & The discussion following Corollary 1 suggests that $\sum_i \hat{v}_{T,i}^{1/2}$ might be much smaller than d $G_{\infty}$. \\
    {\bf Head} & but we should always expect it to be at least a constant,  \\
    {\bf Label} & \texttt{no-rel} \\
\bottomrule
\end{tabular}
\caption{Representative challenging examples during argument relation annotation on AMPERE++.}
\label{tab:appendix-difficult}
\end{table*}

\begin{table*}[t]
\centering

\scalebox{0.95}{
\begin{tikzpicture}
    \node[draw, inner sep=10pt, rounded corners]{
    \begin{tabular}{p{145mm}}
         
    \textbf{General Instruction} \\
         
    In the following studies, you will read a total of 400 peer reviews collected from the ICLR-2018 conference. The annotation is carried out in 20 rounds. In each round, you will independently annotate 20 reviews and upload to the  server. All annotators will meet and discuss the disagreements. Another judge will resolve the cases and add it to the pool of samples for future reference.
    \\
    \\
    \textbf{Annotation Schema} \\
    
    Each review document is already segmented into chunks of argumentative discourse units (ADU), which is the basis for relation annotation. Prior work has provided labels for types of these ADUs: \\
    
    \quad\textsc{Evaluation}: Subjective statements, often containing qualitative judgement. \\
    \quad\textsc{Request}: Statements requesting a course of action. \\
    \quad\textsc{Fact}: Objective information of the paper or commonsense knowledge. \\
    \quad\textsc{Reference}: Citations or URLs. \\
    \quad\textsc{Quote}: Quoations from the paper. \\
    \quad\textsc{Non-Arg}: Non-argumentative statements. \\
    \\
    Please first read the entire review. Then, from the beginning of the document, start annotating support and attack relations.
    We consider a support relation holds from proposition \texttt{A} to proposition \texttt{B} if and only if the validity of \texttt{B} can be undermined without \texttt{A}, or \texttt{A} presents concrete examples to generalize \texttt{B}. For example, ``\textit{It is unclear which hacks are the method generally.}'' is supported by ``\textit{Because the method is only evaluated in one environment.}''.
    
    \\
    We consider an attack relation holds from proposition \texttt{A} to proposition \texttt{B} if and only if \texttt{A} contrasts or questions \texttt{B}'s stance. For example, ``\textit{The authors mentioned that the grammar in general is not context free.}'' is attacked by ``\textit{But the grammar is clearly context-free.}''
    \\
    \\
    Both the support and attack relations can be implicit or explicit. Explicit relations are indicated by discourse markers, whereas implicit relations require inference from the context. For example, ``\textit{In particular, how does the variational posterior change as a result of the hierarchical prior?}'' implicitly supports ``\textit{It's not clear as to why this approach is beneficial}''. Because the question instantiates the ``unclear'' claim regarding the approach. \\
    \\
    \textbf{Special Cases}\\
    Please enforce the following constraints: \\
    \quad 1. The factual propositions (i.e., \textsc{Fact}, \textsc{Reference}, \textsc{Quote}) cannot be supported by any subjective propositions (i.e., \textsc{Evaluation}, \textsc{Request}). \\
    \quad 2. One proposition can support or attack at most one proposition. \\
    \quad 3. Chain support does not need to be explicitly annotated. For instance, if \texttt{A} supports \texttt{B}, \texttt{B} supports \texttt{C}, then \texttt{A} supports \texttt{C} does not need annotation.\\

    \end{tabular}};
\end{tikzpicture}}

\caption{Argumentative relation annotation guideline for AMPERE++.}
\label{tab:anno}
\end{table*}